\documentclass[runningheads]{llncs}
\usepackage[T1]{fontenc}
\usepackage{graphicx}
\usepackage{hyperref}
\usepackage{booktabs} 
\usepackage{amsmath}
\usepackage{cleveref}
\usepackage{float}
\begin{document}
\title{Interpret, prune and distill Donut : \\ towards lightweight VLMs for VQA on documents}
\titlerunning{Donut-MINT : towards leightweight VLMs for document VQA}
% If the paper title is too long for the running head, you can set
% an abbreviated paper title here
%
\author{Adnan Ben Mansour\inst{1,2,3}\orcidID{0009-0002-1891-4903} \and Ayoub Karine\inst{2}\orcidID{0000-0002-9304-4613} \and David Naccache\inst{1,3}\orcidID{0000-0002-8651-6555}}

\authorrunning{A. Ben Mansour et al.}
% First names are abbreviated in the running head.
% If there are more than two authors, 'et al.' is used.
%
\institute{DIENS, École Normale Supérieure of Paris, Paris, France \\ \email{adnan.ben.mansour@ens.fr \footnote{corresponding author}}
\and
Université Paris Cité, LIPADE, F-75006 Paris, France 
\and
Be-ys Research, Paris, France}

\maketitle              % typeset the header of the contribution
\begin{abstract}
Recent advances in Visually-rich Document Understanding rely on large Vision-Language Models like Donut, which perform document-level Visual Question Answering without Optical Character Recognition. Despite their effectiveness, these models are too costly for real-time or resource-constrained applications. We investigate model compression through knowledge distillation, training compact student models from a larger teacher. We leverage mechanistic interpretability to drive student architecture design within this framework. By analyzing internal computations, we identify essential subcomponents to retain, while having a clear view of which subcomponents should be approximated, skipped, or reparametrized based on their function. This approach yields Donut-MINT (Mechanistic Interpretability-based Network Trimming), a pruned Donut variant that reduces inference time and memory usage while maintaining strong performance on DocVQA, a standard benchmark for document Visual Question Answering. Our method reframes compression as circuit discovery, bridging interpretability research and practical Vision-Language Model deployment.

\end{abstract}

\section{Introduction} \label{introduction}
% context
Visually-rich Document Understanding (VrDU) is a key area in multimodal artificial intelligence, with applications including invoice parsing, contract analysis, and flowchart interpretation. A central task is Visual Question Answering (VQA) on documents, where models answer questions about the content of scanned or rendered documents. Traditional approaches often rely on OCR-based pipelines~\cite{li2022ppocrv3attemptsimprovementultra,li2022trocrtransformerbasedopticalcharacter,Xu_2020}, which are brittle and domain-specific. Recently, end-to-end Vision-Language Models (VLMs) that bypass OCR have become increasingly popular.

% sota
Several VLMs have advanced the state of the art in document VQA, including Donut~\cite{kim2022ocrfreedocumentunderstandingtransformer}, Pix2Struct~\cite{lee2023pix2structscreenshotparsingpretraining}, DocformerV2~\cite{appalaraju2023docformerv2localfeaturesdocument}, Qwen2.5-VL~\cite{bai2025qwen25vltechnicalreport} and InternVL3~\cite{zhu2025internvl3exploringadvancedtraining}. These models are generally based on transformer architectures~\cite{vaswani2023attentionneed,dosovitskiy2021imageworth16x16words}, combining a vision backbone with language modeling objectives to process document content holistically.

% critic size
Despite strong performance, these models are computationally expensive and difficult to deploy in real-world environments with latency or resource constraints. This raises a fundamental challenge: how can we compress large VLMs while preserving task accuracy? While not the most powerful model, Donut remains a strong candidate for compression due to its OCR-free formulation and favorable trade-off between size and performance. For instance, the Donut-Hole method~\cite{shaikh2023donutholedonutsparsificationharnessing} compresses Donut via unstructured magnitude-based pruning followed by knowledge distillation~\cite{hinton2015distillingknowledgeneuralnetwork}. In this setting, the pruned model mimics the teacher’s output distributions, recovering lost performance without changing the architecture.

% critic black-box
However, such approaches treat the teacher model as a black box. This lack of transparency limits our ability to guide compression using internal signals. In practice, VLMs used in VrDU remain poorly understood. Existing pruning and distillation methods are mostly blind to model internals, offering no principled way to distinguish essential from redundant components.

% our method
In this work, we ask whether mechanistic interpretability (MI) can help address this limitation. MI seeks to uncover the internal representations and computational mechanisms that drive model behavior. We apply MI tools to Donut’s attention patterns and activations to identify functionally important components and remove others in a targeted fashion. Rather than distilling from a model we do not understand, we leverage internal insights to inform pruning and student design.

% why donut
We focus our study on Donut for several reasons. First, Donut is relatively small, with approximately 250 million parameters and a shallow multimodal decoder of four transformer layers, making systematic analysis feasible. Second, Donut uses explicit cross-attention between visual features and decoder tokens, making its modality fusion interpretable and structurally disentangled. Third, it is trained to copy text spans from the image rather than generate free-form answers, which makes attribution of internal mechanisms more tractable. While our study centers on Donut, we view it as a proof of concept for a broader class of VLMs.

\begin{figure}[h]
	\vskip 0.2in
	\begin{center}
		\centerline{\includegraphics[width=0.6\textwidth]{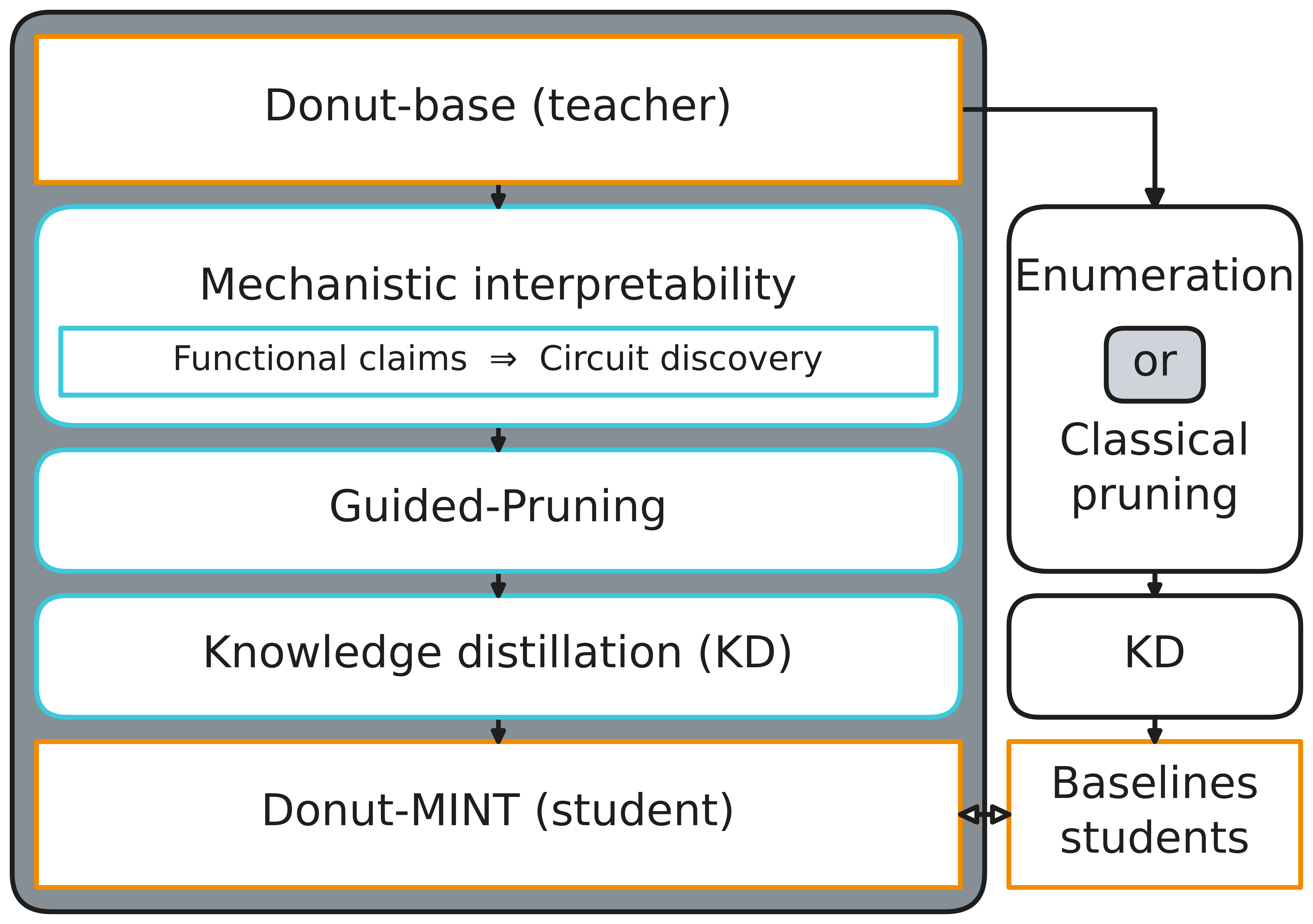}}
		\caption{Our proposed method relies on the three stages highlighted in blue: MI-based importance estimation, pruning, and knowledge distillation. We also pruned Donut-base with other SOTA techniques to allow for a fair comparison.}
		\label{fig: flowchart}
	\end{center}
	\vskip -0.2in
\end{figure}

% conclusion
Our experiments show that our pruned model, Donut-MINT (Mechanistic Interpretability-guided Network Trimming), achieves significant reductions in both parameters and computation, while closely matching the teacher model’s accuracy on DocVQA after distillation. An overview of our approach is provided in Figure~\ref{fig: flowchart}, highlighting the main stages of our method and how they compare to other baselines.

% contributions

\textbf{Our contributions} are threefold:
(1) We provide a high-level interpretation of how an end-to-end VLM solves VQA on documents, identifying the computational patterns and representations most critical to the task.
(2) We demonstrate that these insights can inform the design of a student model, outperforming baseline pruning or blind distillation strategies.
(3) We introduce Donut-MINT, a lightweight model that achieves competitive performance on the DocVQA dataset~\cite{mathew2021docvqadatasetvqadocument}, thereby validating our methodology and illustrating how interpretability can drive principled model compression.

% outline
The rest of this paper is structured as follows. \cref{sec: related work} surveys prior work on mechanistic interpretability, model pruning, and knowledge distillation. \Cref{sec: method} presents our proposed framework, details the Donut architecture, and describes the tools used to derive efficient student models. \cref{sec: results} provides our interpretability claims and reports our experiments on the DocVQA dataset. \cref{sec: conclusion} outlines implications, limitations, and future directions.

\section{Related work} \label{sec: related work}
\subsection{Mechanistic Interpretability}
Mechanistic interpretability (MI) seeks to reverse-engineer internal computations in neural networks, departing from traditional interpretability by focusing on the mechanisms themselves. As described by \cite{rai2025practicalreviewmechanisticinterpretability}, MI is structured around three components: \textit{features}, \textit{circuits}, and \textit{universality}. Features are human-interpretable properties encoded in activations; circuits are the paths connecting and processing these features; universality concerns the redundancy of such components across diverse models.

\subsubsection{Core methods}
According to \cite{bereska2024mechanisticinterpretabilityaisafety}, MI methods fall into two categories: intervention-based and observational.

\textbf{Intervention-based methods} operate through three runs: a clean run, a corrupted run, and a patched run. The core idea is to selectively replace activations from the clean run with those from the corrupted run, or conversely (in denoising-based variants), to restore clean behavior by overwriting corrupted activations. \cite{olsson2022incontextlearninginductionheads} apply this to attention heads using ablation. Corruption can involve zeroing activations, but alternatives have been proposed, such as adding random noise \cite{rai2024investigationneuronactivationunified} or replacing activations with their mean value over a small dataset \cite{wang2022interpretabilitywildcircuitindirect}. Activation patching is closely related to causal mediation analysis \cite{NEURIPS2020_92650b2e} and is referred to as causal tracing in \cite{meng2023locatingeditingfactualassociations}. In some works \cite{meng2023locatingeditingfactualassociations,hanna2023doesgpt2computegreaterthan}, the corruption procedure involves resampling: the clean and corrupt runs are generated from different inputs, rather than modifying a single input. Attribution patching \cite{nanda2023attributionpatching} replaces iterative patching with gradient-based estimates, accelerating discovery of circuits.

\textbf{Observational approaches} require no model alteration. Vocabulary projection methods (e.g., logit lens \cite{nostalgebraist2020logitlens}, attention lens \cite{sakarvadia2023attentionlenstoolmechanistically}) decode activations via the unembedding matrix. Probing techniques \cite{gurnee2023findingneuronshaystackcase} and sparse autoencoders \cite{burns2024discoveringlatentknowledgelanguage} train auxiliary models to recover encoded information, either via supervision or unsupervised reconstruction.

\subsubsection{MI for VLMs}
MI has progressed rapidly for LLMs, while VLMs remain underexplored. Recent efforts extend LLM techniques to the multimodal setting. \cite{lin2025surveymechanisticinterpretabilitymultimodal} distinguish between VLM adaptations of LLM tools and VLM-specific methods. Causal tracing was applied to vision encoders \cite{palit2023visionlanguagemechanisticinterpretabilitycausal} and full VLMs \cite{basu2024understandinginformationstoragetransfer}, identifying locations of stored knowledge. In VQA, \cite{yu2025understandingmultimodalllmsmechanistic} study visual hallucinations in LLaVA \cite{liu2023visualinstructiontuning}.

Dedicated tools include adaptations of Symmetric Token Replacement to the image modality \cite{golovanevsky2025vlmsnoticemechanisticinterpretability}, text decodings of image embeddings \cite{gandelsman2024interpretingclipsimagerepresentation}, and neuron-level analyses \cite{bau2017networkdissectionquantifyinginterpretability,bai2025interpretingneuronsdeepvision}. Interpretability techniques leveraging cross-attention exist in generative models \cite{hertz2022prompttopromptimageeditingcross,basu2023localizingeditingknowledgetexttoimage}, but have not been adapted to VLMs. Crucially, to the best of our knowledge, no prior work addresses interpretability in the specific context of Visual Question Answering on documents. 

\subsection{Compression Techniques for Transformer-based Models}
Compression of transformer-based models, which form the backbone of most VLMs, typically falls into three categories: pruning, knowledge distillation, and quantization. Since quantization alters weight precision without changing model structure, it is largely orthogonal and not considered here. We focus on methods that reshape the model or its training: structured pruning, knowledge distillation, and their combination. We conclude by reviewing how these approaches extend to vision-language models.

\subsubsection{Structured pruning and knowledge-distillation}
Structured pruning removes entire components (layers, heads, neuron groups) to reduce size and computation. \cite{zhang2024finercutfinergrainedinterpretablelayer} prune feed-forward and attention blocks, \cite{hu2025faspfastaccuratestructured} remove submatrix structures, and \cite{NEURIPS2024_c1c44e46} use gradient-guided strategies. The lottery ticket hypothesis \cite{frankle2019lotterytickethypothesisfinding} suggests pruned subnetworks can perform well when reinitialized properly. However, pruning often reduces performance, necessitating retraining or fine-tuning.

Knowledge distillation \cite{hinton2015distillingknowledgeneuralnetwork} compresses models by aligning a compact student with a larger teacher. Techniques include output imitation, intermediate feature matching \cite{romero2015fitnetshintsdeepnets}, attention alignment \cite{zagoruyko2017payingattentionattentionimproving}, and structural preservation \cite{tung2019similaritypreservingknowledgedistillation}. These improve generalization while enabling smaller models.

Combining pruning and distillation enhances performance recovery after pruning. \cite{muralidharan2024compactlanguagemodelspruning} use this strategy to compress LLMs without compromising downstream accuracy. The teacher is the original model; the student is the pruned one retrained to mimic the teacher’s outputs.

\subsubsection{VLM compression for VrDU}
The closest attempt at combining pruning and distillation in VLMs for VrDU tasks, although applied to document reading and key information extraction rather than VQA, is Donut-Hole \cite{shaikh2023donutholedonutsparsificationharnessing}. However, their approach primarily focuses on reducing the parameter count through unstructured pruning \cite{hanIMP,zafrir2021pruneallsparsepretrained,back2023magnitudeattentionbaseddynamicpruning,sun2024simpleeffectivepruningapproach}, explicitly setting a subset of weights to zero based on their magnitude and forcing these weights to remain zero during retraining with KD. While this produces a model with fewer active weights, achieving actual speedups typically requires specialized hardware for sparse inference.

In another line of research, token pruning \cite{wang2024smarttrimadaptivetokensattention,ye2024atpllavaadaptivetokenpruning,yang2025intermediatestatesexplainingvisual} has been explored, with the goal of discarding tokens in the cross-attention mechanism based on their importance or redundancy in the input image. However, this approach is inherently challenging to apply in VrDU, as most of the patches in document images are crucial for understanding and contain valuable information.

\section{Method} \label{sec: method}
\subsection{Our framework} \label{subsec: our framework}
\Cref{fig: flowchart} outlines our framework. Starting from Donut-base as the teacher model, we apply mechanistic interpretability techniques to analyze its behavior on document-based questions, yielding insights formalized in \cref{subsec: interp insights}. These guide a structured pruning process to obtain a distilled student model. For comparison, we also perform an exhaustive search over candidate student architectures using both structured and unstructured coarse-grained pruning strategies, following \cite{shaikh2023donutholedonutsparsificationharnessing}. All student models are trained via knowledge distillation from the teacher. Our MI-guided pruning proceeds in two stages: first, we prune sub-layers (self-attentions, cross-attentions, and feed-forwards) from the multimodal decoder; second, we apply fine-grained pruning to individual attention heads within the remaining self and cross-attention layers.

\subsection{Donut overview}
We conduct all experiments on Donut \cite{kim2022ocrfreedocumentunderstandingtransformer}, used as our teacher model. It combines a Swin-B vision encoder \cite{liu2021swintransformerhierarchicalvision} with a multimodal decoder composed of the first four layers of mBART \cite{liu2020multilingualdenoisingpretrainingneural}. Donut is a multilingual model pre-trained on 2M synthetic and 11M scanned document images. We use the donut-base-finetuned-docvqa checkpoint, already fine-tuned on DocVQA.

\begin{figure}[h]
	\vskip 0.2in
	\begin{center}
		\centerline{\includegraphics[width=0.6\textwidth]{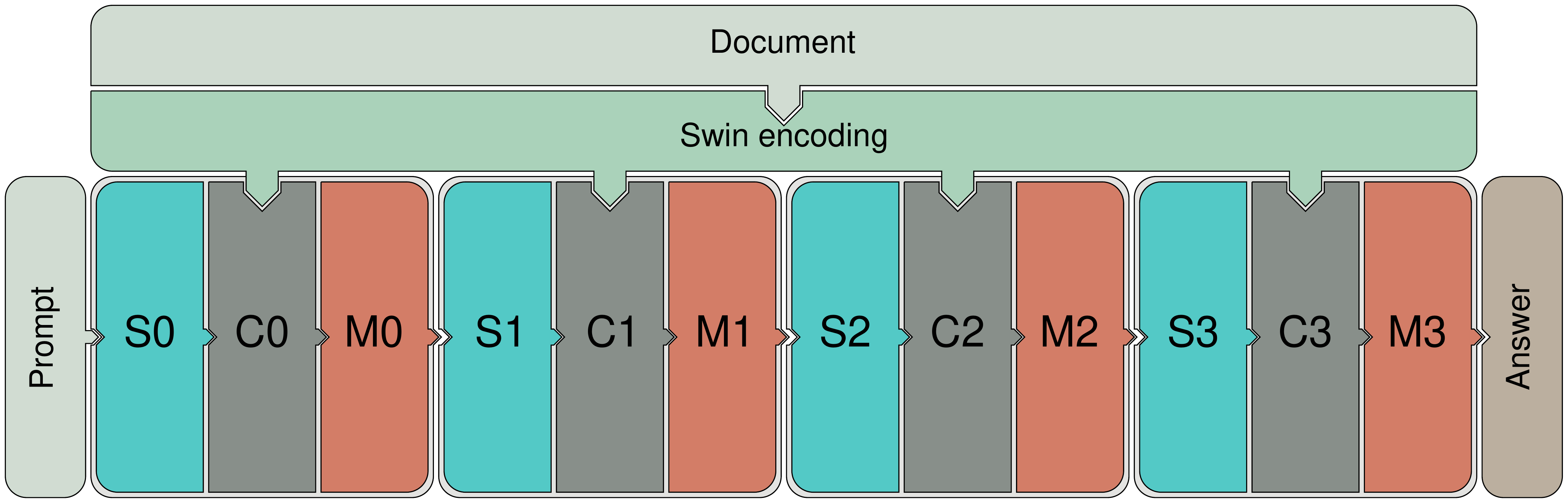}}
		\caption{Architecture overview of Donut-base. The model comprises a visual encoder and a decoder. The decoder consists of four layers, each layer $\ell \in \{0,1,2,3\}$ divided into three sub-layers: self-attention \texttt{S$\ell$}, cross-attention \texttt{C$\ell$} bridging modalities, and feed-forward networks \texttt{M$\ell$}.}
		\label{fig: donutbigpicture}
	\end{center}
	\vskip -0.2in
\end{figure}

\Cref{fig: donutbigpicture} illustrates Donut’s architecture, showing how the document image and question prompt are processed. We denote sub-layers of the four decoder layers as \texttt{S0}–\texttt{S3} (self-attentions), \texttt{C0}–\texttt{C3} (cross-attentions), and \texttt{M0}–\texttt{M3} (feed-forwards). Attention heads are indexed as in \texttt{C3.H11}, referring to the 12th head in the final cross-attention. Each sub-layer has associated activations (e.g., \texttt{C3.input}, \texttt{C3.output}), defined as needed.

\subsection{Mechanistic Interpretability}
Mechanistic interpretability (MI) provides both an investigative lens into model internals and a foundation for structured pruning. We use MI to dissect the internal computation of Donut-base in the context of single-page document visual question answering (VQA), with the goal of identifying which components—layers, heads, or neurons—are responsible for key aspects of reasoning and representation. These insights directly inform our pruning strategy, allowing us to preserve only components essential to task performance.

To evaluate the functional relevance of model components, we employ activation patching in a teacher-forcing setting. In this setup, the model receives the full input sequence—including the question and the ground-truth answer—enabling us to compute token-level predictions for all decoding steps in a single forward pass. We monitor changes in perplexity, defined as the exponential of the cross-entropy loss, to assess the impact of patching. We also extend logit lens analysis to decoder attention heads (when dimensionality permits) for interpretability at intermediate layers.

Additionally, we apply tailored statistical measures to estimate the importance of attention heads. For self-attention (SA), we compute entropy to measure how concentrated or diffuse each head's attention is across the sequence. High entropy indicates that a token is attending to many others with similar weight, diluting the information it receives. This diffuse attention often signals that the head is noisy or uninformative. In contrast, for cross-attention (CA), which links decoder tokens to Swin-encoded image patches, we use variance. Salient cross-attention heads typically focus sharply on a few high-response patches, while low-variance heads distribute attention more evenly, often reflecting low task relevance. Due to spatial redundancy in Swin features, variance is a more robust proxy than entropy in this context.

\subsection{Guided Pruning} \label{subsec: guidedpruning}
We translate the above mechanistic insights into a two-stage guided pruning strategy, resulting in two compact student models: Donut-MINT$_{31\%}$ and Donut-MINT$_{7\%}$.

In the first stage, we prune at the sub-layer level. Mechanistic analysis identifies the initial feed-forward sub-layer (\texttt{M0}) as dispensable, and we show its function can be absorbed by adjusting the tokenizer or embedding layer. Building on this, we construct a student model retaining only $31\%$ of the original parameters (Donut-MINT$_{31\%}$), and further reduce it to $18\%$ by removing \texttt{M0}. Sub-layer pruning in both cases preserves the minimal circuits identified as necessary to support the full range of document VQA subtasks.

In the second stage, we apply fine-grained pruning to attention heads based on the statistical indicators described above. We remove self-attention heads with high entropy and cross-attention heads with low variance, treating them as proxies for functional irrelevance. Starting from the $18\%$ variant, we eliminate these heads and selectively reintroduce others critical for specific sub-tasks, resulting in Donut-MINT$_{7\%}$, a model with just 7\% of the original parameters but competitive performance.    

\subsection{Knowledge distillation}
All student models are initialized from the pretrained weights of Donut-base and retrained using knowledge distillation on the DocVQA training dataset. We follow the standard setup introduced by \cite{hinton2015distillingknowledgeneuralnetwork}, which combines supervision from the ground-truth label with soft targets produced by the teacher model. The total loss for student training is defined as:

\begin{equation}
\mathcal{L}_{\textrm{total}} = \mathcal{L}_{\textrm{CE}} + \alpha \mathcal{L}_{\textrm{KD}}
\end{equation}

Here, $\mathcal{L}_{\text{CE}}$ is the cross-entropy loss between the student's logits and the ground-truth labels. The $\mathcal{L}_{\text{KD}}$ term represents the softened Kullback-Leibler (KL) divergence between the teacher’s and student’s output distributions, computed from logits scaled by a temperature $T$. During training, the teacher model is kept frozen, as are the encoder and embedding layers of the student; only the student decoder is updated. The weighting factor $\alpha$ controls the trade-off between the ground-truth supervision and the distillation signal : it is chosen such that both loss terms contribute with similar magnitudes. 

\section{Results} \label{sec: results}
\subsection{Setup}
We use the official Donut-base model pretrained on DocVQA as our teacher, avoiding costly pretraining while representing a realistic deployment target. Student models are retrained on the full DocVQA training set using our distillation pipeline. All experiments run on two NVIDIA L40S GPUs.

We evaluate on DocVQA \cite{mathew2021docvqadatasetvqadocument}, a benchmark of 50,000 QA pairs over 12,000+ visually diverse documents. We split its original validation set into three parts: one for interpretability analysis (20\%), one for student model selection (20\%), and one for post-pruning evaluation (60\%). Since test annotations are private, final accuracy is reported via the official evaluation server \footnote{\url{https://rrc.cvc.uab.es/?ch=17}, accessed in April 2025.}.

Performance is measured using ANLS \cite{biten2019scenetextvisualquestion}, which computes normalized edit similarity between predictions and references with a cutoff at $\tau = 0.5$. This balances tolerance to minor typos with sensitivity to semantic correctness.

To assess the efficiency of the compressed model, we adopt two metrics at a given performance level: the number of parameters and the inference cost measured in floating-point operations (FLOPs). \Cref{fig: flops} reports the value of these metrics for the different sub-layers of Donut-base, highlighting the primary computational bottlenecks for each metric. Parameter count is a coarse measure of model size and storage requirements, but it fails to capture actual computational cost, especially in autoregressive generation tasks where the same parameters may be reused across multiple steps. FLOPs, by contrast, estimate the number of arithmetic operations required for a full forward pass and provide a hardware-agnostic proxy for runtime and energy consumption.

\begin{table}[h]
	\caption{Number of parameters and FLOPs required by Donut-base components to generate $m=35$ new tokens given $n=42$ prompt tokens. G and T refer to $10^9$ (giga) and $10^{12}$ (tera) FLOPs, respectively, while M stands for $10^6$ parameters.} \label{fig: flops}
	\vskip 0.15in
	\begin{center}
		\begin{small}
			\begin{sc}
				\begin{tabular}{lrr}
					\toprule
					Module & Parameters & FLOPs  \\
                        \midrule
                        \midrule
                        Donut-base & 257M & 2.99T \\
                        \midrule
                        \qquad Encoder & 75M & 1.30T \\
					   \qquad Embedding & 59M & 0 \\
					   \qquad Unembedding & 59M & 117.17G \\
                        \midrule
                        \qquad Decoder & 64M & 1.57T \\
					\qquad \qquad Self-Attention $\ell$ & 4M & 8.58G \\
					\qquad \qquad Cross-Attention $\ell$ & 4M & 366.11G \\
					\qquad \qquad Feed-Forward $\ell$ & 8M & 16.69G \\
					\bottomrule
				\end{tabular}
			\end{sc}
		\end{small}
	\end{center}
	\vskip -0.1in
\end{table}

While the encoder contributes most of the FLOPs in a single forward pass, we focus our analysis on the decoder for both conceptual and practical reasons. Structurally, the decoder is multimodal and must combine visual and textual information, making it more likely to implement complex and intertwined sub-tasks. This makes it a natural target for mechanistic interpretability and pruning. On the practical side, decoder FLOPs grow with both the context length and the number of generated tokens. Since document complexity often correlates with context size, the decoder’s computational burden tends to dominate in realistic deployment scenarios.

\subsection{Interpretability insights} \label{subsec: interp insights}
\subsubsection{Transcription}
\begin{proposition}
	The task of converting pixel values into discrete tokens is entirely done in the visual encoder up to a linear transformation. \texttt{C3} puts attention on patches it wants to read, and heads \texttt{C3.H4}, \texttt{C3.H7}, \texttt{C3.H8}, \texttt{C3.H9}, \texttt{C3.H10}, and \texttt{C3.H11} retrieve the token at that position. 
\end{proposition}

\begin{figure}[H]
    \vskip 0.2in
    \begin{center}
        \centerline{\includegraphics[width=0.7\columnwidth]{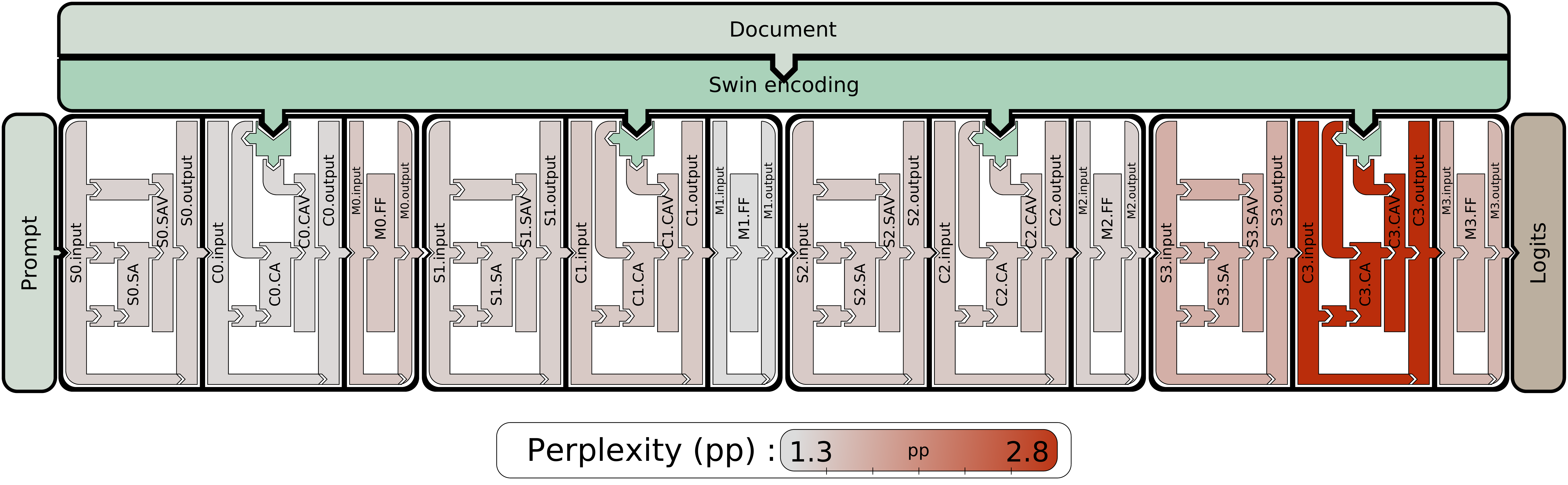}}
        \caption{Impact of skipping decoder sub-layers in Donut-base measured by perplexity. Each sub-layer of the decoder is color-coded according to the increase in perplexity observed when that sub-layer is skipped, highlighting which components are critical to model performance.}
        \label{fig: skipping}
    \end{center}
    \vskip -0.2in
\end{figure}

Skipping analysis identifies \texttt{C3} as the most crucial sub-layer (\cref{fig: skipping}) and Logit-lens confirms that it outputs the desired text. By directly projecting encoder embeddings through \texttt{lm\_head $\circ$ C3.out\_proj $\circ$ C3.v\_proj}, we can fully transcribe the image patch-wise.

\begin{figure}[H]
    \centering

    \hfill
    \includegraphics[width=0.44\textwidth,keepaspectratio]{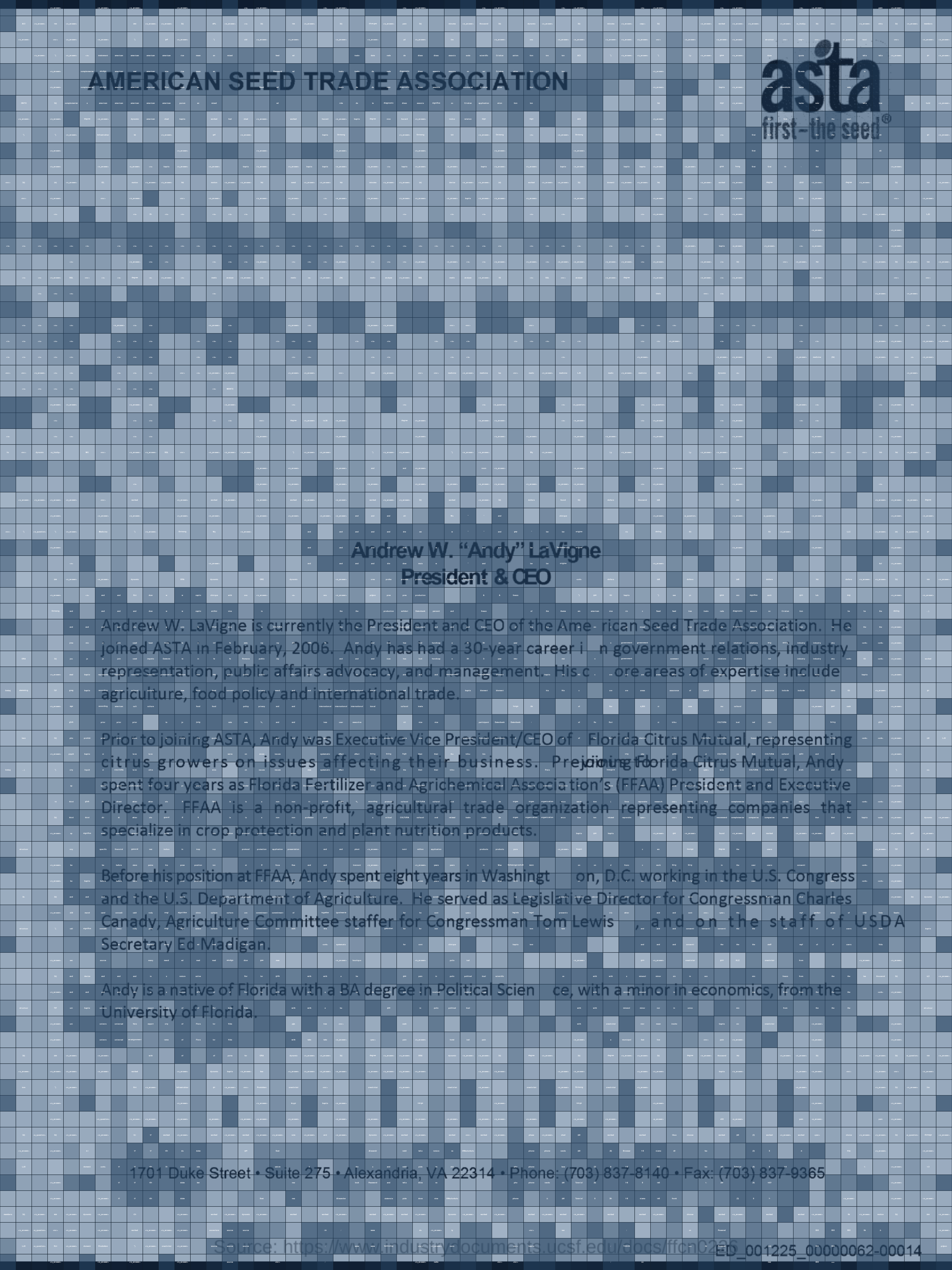}
    \hfill
    \includegraphics[width=0.44\textwidth,keepaspectratio]{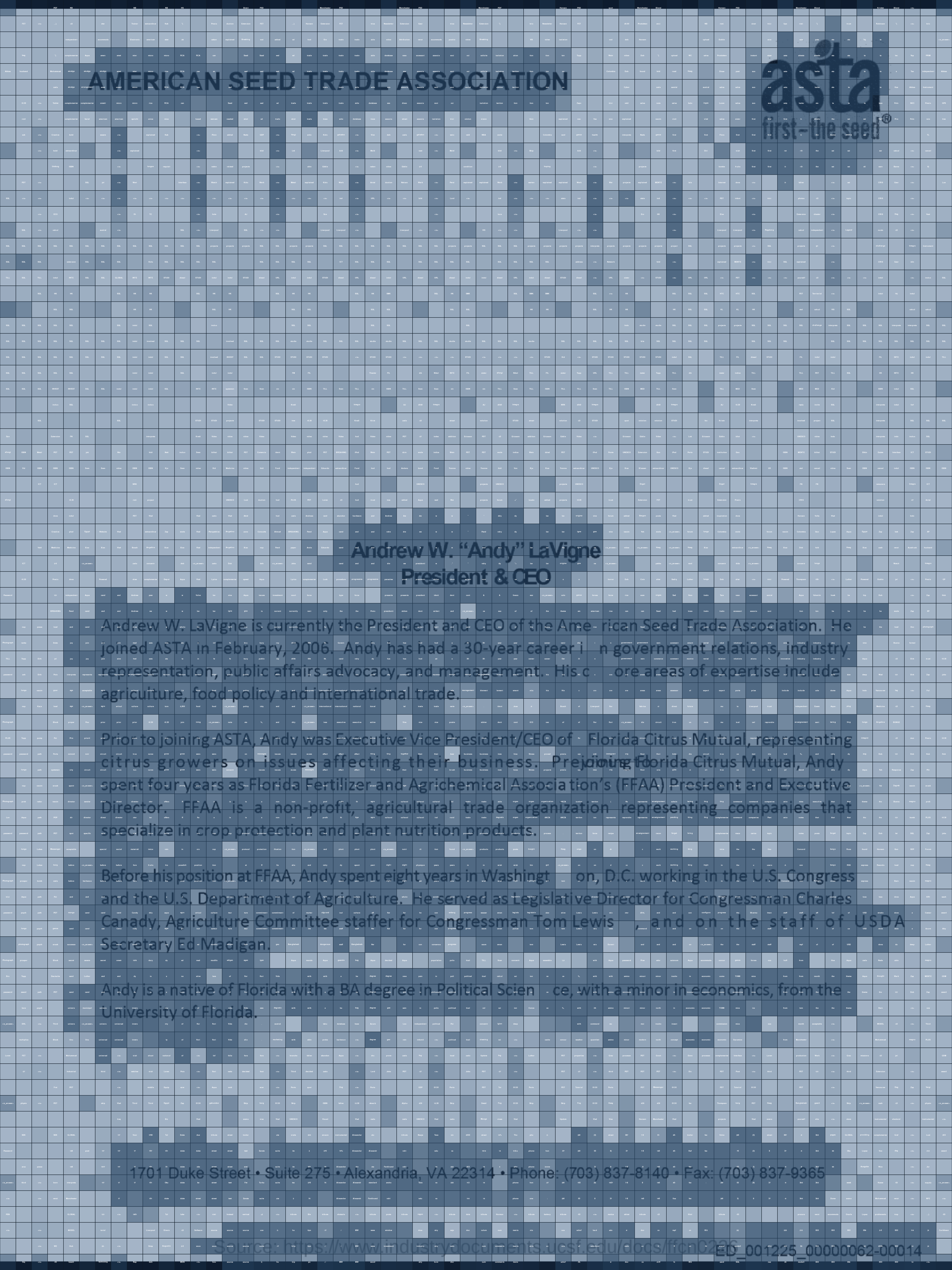} 
    \hfill
    {}
    
    \vspace{0.5em}
    
    \includegraphics[width=0.98\textwidth,keepaspectratio]{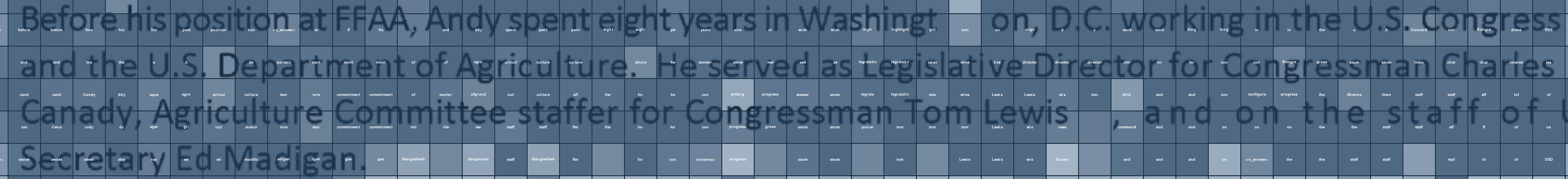}

    \caption{(left) Projection of patch visual encoding through Swin and the linear transformation \texttt{lm\_head $\circ$ C3.out\_proj $\circ$ C3.v\_proj} from the decoder into tokens. We colored in blue patches with a low entropy on the token distribution, and overlayed the most likely token for each. (right) We removed least important heads, leading to a clear improvement over text transcription fidelity. (bottom) This is a zoomed version to improve the lisibility of the text.}
    \label{fig: transproj}
\end{figure}

To identify the most relevant attention heads for this task, we select those with the lowest cross-attention map variance. We then project into the vocabulary space using only these heads, and we highlight image patches with low entropy in that space to better visualize which regions contribute most confidently to the transcription (\cref{fig: transproj}).

\begin{proposition}
	\texttt{M3} sole purpose is to remove the residual stream from \texttt{C3} which was necessary to construct the cross-attention maps. 
\end{proposition}

\begin{figure}[H]
  \vskip 0.2in
  \begin{center}
    \begin{minipage}{0.48\linewidth}
      \centering
      \includegraphics[width=\linewidth]{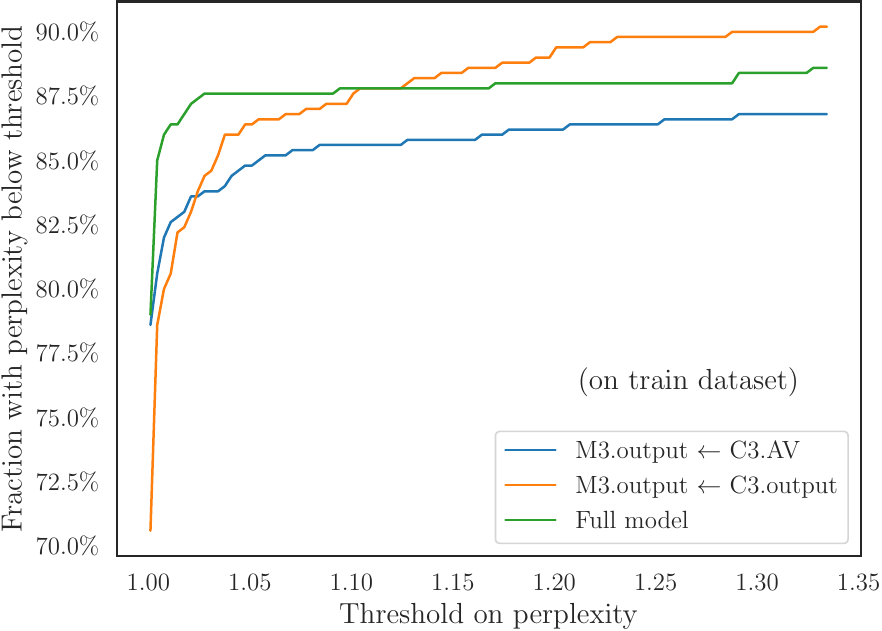}
    \end{minipage}
    \hfill
    \begin{minipage}{0.48\linewidth}
      \centering
      \includegraphics[width=\linewidth]{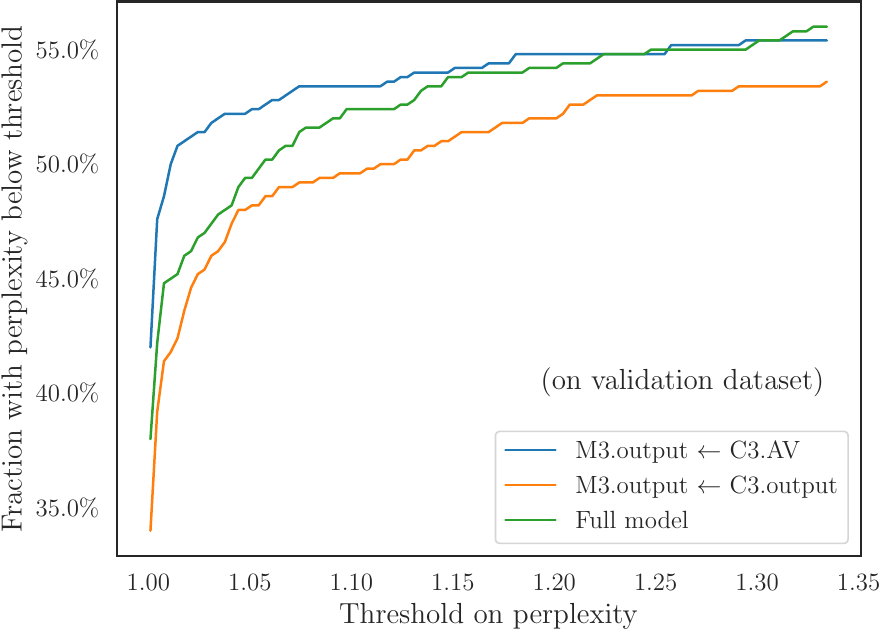}
    \end{minipage}
    \caption{Accuracy versus perplexity threshold for the first generated token of the \texttt{M3} sub-layer after activation patching. Results are shown on both train and validation splits, plotting the fraction of samples with perplexity below each threshold under three conditions: skipping \texttt{M3} entirely, patching \texttt{M3} with activations from the last cross-attention times value (\texttt{C3.AV}), and the clean run.}
    \label{fig:m3}
  \end{center}
  \vskip -0.2in
\end{figure}

Since we know that \texttt{C3} already has the answer and Donut-base only outputs text present in the document, keeping \texttt{M3} becomes questionable. We found that, in fact, it removes the residual stream from \texttt{C3} thanks to logit-lens. These findings are also supported by skipping, cross-attention inspection, and activation patching. Removing \texttt{M3}, or equivalently removing \texttt{C3}’s residual, yields near-identical perplexity scores (\cref{fig:m3}), suggesting \texttt{M3} acts as a cleaner rather than a transformer.

\subsubsection{Case-sensitivity}
\begin{proposition}
	The first two sub-layers, \texttt{S0} and \texttt{C0}, transform token embeddings into other token-like embeddings, primarily to align special tokens with those used during pretraining. Feed-forward \texttt{M0} maps tokens to the representation of their lowercase form, effectively neutralizing case distinctions in the embedding space.
\end{proposition}

\begin{figure}[H]
	\vskip 0.2in
	\begin{center}
		\centerline{\includegraphics[width=0.7\textwidth]{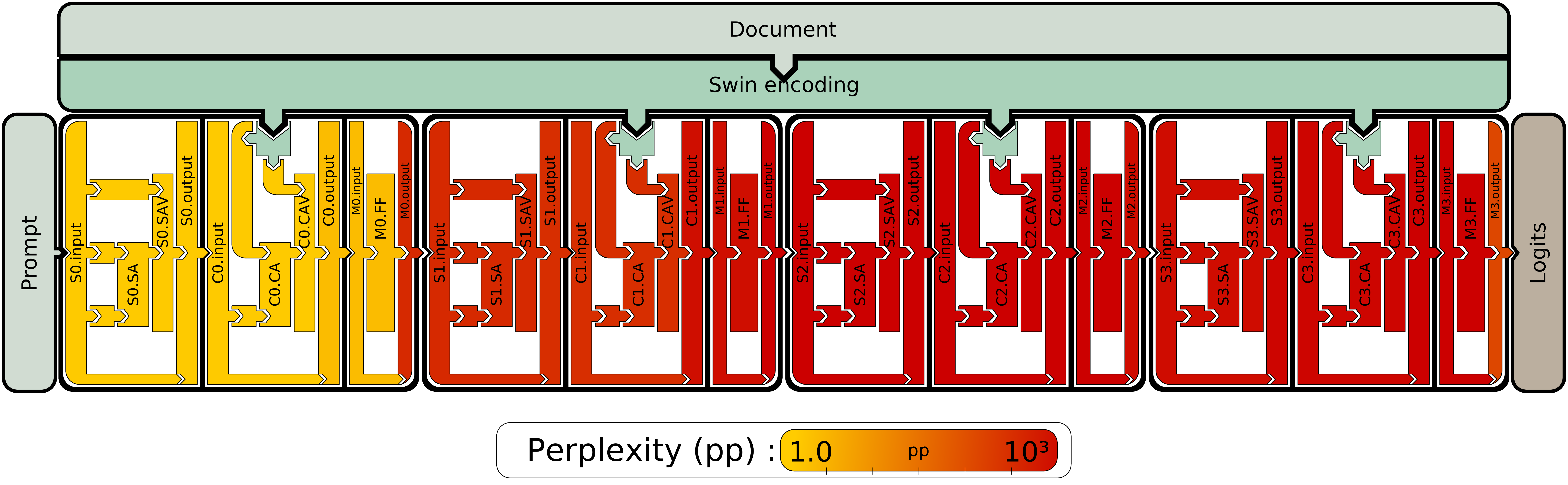}}
		\caption{Impact of \textit{token reprojection} on decoder sub-layers in Donut-base measured by perplexity. Each sub-layer of the decoder is color-coded according to the increase in perplexity observed when that sub-layer activations are replaced with their reprojected embeddings.}
		\label{fig: rect}
	\end{center}
	\vskip -0.2in
\end{figure}

Using logit-lens and a novel technique we call \emph{token reprojection}, we show that early decoder layers operate within a token embedding subspace. Given activation $x$, we reproject as $x' = \operatorname{embedding}(\arg\max(\operatorname{lm\_head}(x)))$, collapsing $x$ to its closest interpretable token. Applying this before \texttt{M0} retains output quality, suggesting that \texttt{M0} enforces lowercasing while preserving semantic identity (\cref{fig: rect}).

Logit-lens is sufficient to see that \texttt{M0} converts question representation into lowercase. We introduce a simple technique called \textit{token reprojection}, which allows us to remove a contiguous block of early decoder layers without performance degradation. Given a hidden activation $x$ at any layer, we compute $x' = \operatorname{embedding}\left( \arg\max \left( \operatorname{lm\_head}(x)\right) \right)$, effectively projecting $x$ back onto the vocabulary embedding manifold. This operation enforces a discrete constraint on the representation, collapsing it to the nearest interpretable token embedding. Remarkably, we find that applying token reprojection before the first feed-forward layer (\texttt{M0}) yields outputs nearly indistinguishable from the original model. This indicates that all intervening layers, including \texttt{M0}, operate within or close to the embedding subspace, and can therefore be bypassed or replaced by a simple re-embedding. By projecting activations onto discrete token embeddings, token reprojection tests whether additional information in other dimensions is necessary, enabling aggressive layer pruning without loss of performance.

\subsubsection{Keyword-based Questions}
\begin{proposition}
Keyword-based questions refer to those targeting a generic field such as "date", "title", or "name". In these cases, sub-layers {S0, S1, S2} propagate information from the keyword token to the special \texttt{<s\_answer>} token, transforming its representation into a semantic query. This query guides the model to attend to any instance of the targeted concept within the document.
\end{proposition}

We show the attention flow from the keyword token to \texttt{<s\_answer>} by analyzing the self-attentions. The keyword’s activation is transformed into a query vector that directs attention to all instances of the relevant concept, such as dates. Furthermore, we identify the minimal circuit capable of replicating this activation-to-query transformation. Detailed analysis can be found in \cref{subsec: keyword}.

We first demonstrate that the query vector constructed in \texttt{C3} is semantically grounded: for example, when prompted with “What is the date?”, the corresponding cross-attention consistently targets date-related regions in the image. Visualizing the self-attention graph from the input tokens to \texttt{<s\_answer>} reveals that information propagates from the keyword to the final token, which subsequently produces the query. Notably, this query appears to be independent of the visual input.

To further isolate the mechanism generating the query, we explored all $2^5$ possible signal pathways from the embedding of the keyword token (e.g., \texttt{"date"}) to the query vector in the final cross-attention sub-layer. We considered two types of signal propagation through each sub-layer: (1) residual connections, and (2) either the self-attention value/output projections or, in the case of feed-forward layers, the linear transformations. This exhaustive search revealed that the most effective circuit consists solely of the final self-attention sub-layer.

\subsubsection{Contextualized Questions}
\begin{proposition}
In structured documents such as forms or tables, some questions query values associated with visible keys (e.g., "What is the place of birth of this person?"). The model attends to the key, extracts its positional embedding, and uses this information to form a query that locates the corresponding value—typically by attending to content on the same line or visual axis.
\end{proposition}

We show that the query at \texttt{C3.query} primarily encodes positional information. This suggests that the model first associates the key in the prompt with its visual counterpart in the document, computes its position, and then retrieves the corresponding content. This task requires at least one cross-attention sub-layer to read the key, and a second to fetch the value.

Using cross-attention maps, we confirm that \texttt{C1}, \texttt{C2}, and \texttt{C3} attend to image patches associated with the prompt. On a sanitized synthetic dataset, activation patching confirms that the query formed at \texttt{C3} retrieves content located along the same visual axis as the key.

\begin{figure}[h]
	\vskip 0.2in
	\begin{center}
		\centerline{\includegraphics[width=0.6\columnwidth]{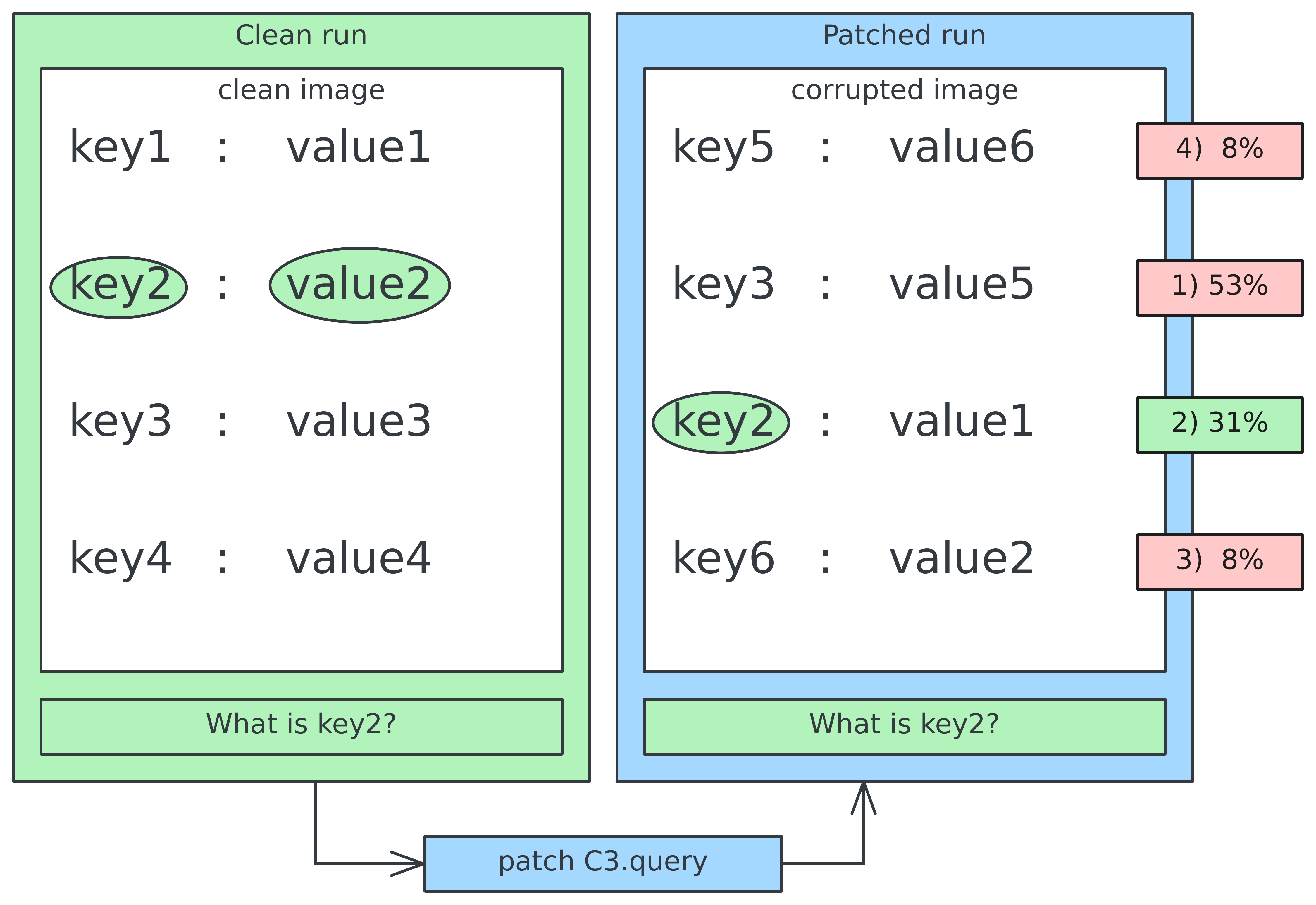}}
		\caption{Evaluating retrieval behavior of the \texttt{C3.query} vector under activation patching in a synthetic key-value setup.
The figure shows two images: the original (clean) and the corrupted version where key-value pairs are shuffled. We patch \texttt{C3.query} and classify the model’s output tokens according to four hypotheses: retrieving the value at the original position (positional retrieval), retrieving the updated correct value (semantic adaptation), retrieving the previously memorized value (value memorization), or producing an unrelated token (failure). Accuracy for each category is displayed in boxes, illustrating the balance between positional bias and semantic adaptation.}
		\label{fig: imagepatching}
	\end{center}
	\vskip -0.2in
\end{figure}

To further analyze how \texttt{C3.query} retrieves content, we build a controlled synthetic setup: key-value pairs drawn from finite sets $K$ and $V$ are placed on the image in the format \texttt{key: value}. The model is prompted with "What is \texttt{key}?", and we evaluate its ability to retrieve the corresponding value. After confirming correct behavior on the clean image, we generate a patched version where values are reassigned randomly to different keys, preserving positions. This setup is illustrated in \cref{fig: imagepatching}.

We evaluate four hypotheses regarding retrieval:

\begin{enumerate}
	\item \textbf{Positional retrieval}: \texttt{C3.query} encodes a fixed spatial position and retrieves the content originally located there, regardless of content changes.
	\item \textbf{Semantic adaptation}: \texttt{C3.query} dynamically adapts and retrieves the correct updated value for the key.
	\item \textbf{Value memorization}: \texttt{C3.query} outputs a memorized token, independent of the prompt or position.
	\item \textbf{Failure mode}: the model produces unrelated tokens, failing to retrieve any relevant information.
\end{enumerate}

When $K = V$ are color names, the model retrieves the value at the original position in 53\% of cases (hypothesis 1), correctly adapts to the new value in 31\% (hypothesis 2), outputs the previously correct value in 8\% (hypothesis 3), and fails in 8\% (hypothesis 4). This reflects a strong positional bias, with moderate adaptability.

Replacing colors with common nouns leads to decreased performance: positional retrieval dominates at 40.4\%, adaptation drops to 26.2\%, memorization accounts for 3.7\%, and failures rise to 29.7\%. This suggests greater ambiguity or weaker priors in these tokens.

When using spelled-out numbers, we observe a more balanced behavior: 37.2\% positional retrieval, 44.6\% semantic adaptation, and 9.1\% each for memorization and failure. This indicates that numerical tokens may support more flexible and robust grounding.

\subsubsection{Generating all the answer}
To continue generating tokens autoregressively, the model leverages the same mechanisms used for answering contextualized questions, since the previously retrieved value now becomes part of the new prompt. Thus, the reading circuit generalizes to multi-token sequences.

\subsection{Comparison with state of the art}
In \cref{fig: table}, we compare our mechanistic interpretability-guided pruning approach to various state-of-the-art (SOTA) techniques for reducing model size and computational cost. We consider both coarse-grained and fine-grained pruning strategies, including structured and unstructured baselines, and report their impact on ANLS, parameter count, and FLOPs.

\textbf{Coarse-grained pruning} removes entire sub-layers in the decoder, targeting both parameter and FLOP reductions while preserving core functionality. Our goal is to reduce the number of parameters to approximately 40\% and FLOPs by half. We benchmark our MI-guided variant, Donut-MINT$_{31\%}$, against the top three students obtained via brute-force enumeration under similar constraints (24M parameters, 49\% FLOPs). We name them Donut-T1, Donut-T2 and Donut-T3. All models, including Donut-MINT$_{31\%}$, are retrained through knowledge distillation to ensure fairness. As shown in \cref{fig: table}, our method consistently outperforms enumeration-based students in ANLS (50\% vs.\ 22--28\%), indicating that MI can more effectively identify critical substructures.

\textbf{Fine-grained pruning} further reduces the model by removing individual attention heads. This yields Donut-MINT$_{7\%}$, produced by applying MI-guided structured pruning to eliminate the least functionally relevant heads, with a target of 7\% of the original decoder parameters. We compare this to two strong baselines. First, the structured pruning method from \cite{zhang2024finercutfinergrainedinterpretablelayer}, which removes components based on saliency metrics. Second, the unstructured pruning approach Wanda \cite{sun2024simpleeffectivepruningapproach}, which prunes weights using magnitude and activation norm heuristics. All three models operate under similar parameter budgets (7\%) and comparable FLOPs (Donut-MINT: 13\%, Structured: 12\%, Unstructured: 100\%). As seen in \cref{fig: table}, Donut-MINT$_{7\%}$ achieves the highest ANLS (51\%) despite the aggressive pruning, outperforming both alternatives (50\% and 46\%).

\begin{table*}[t]
    \caption{Performance and efficiency comparison of student models on DocVQA under coarse-grained and fine-grained pruning constraints.} \label{fig: table}
    \vskip 0.15in
    \begin{center}
        \begin{small}
            \begin{sc}
                \begin{tabular}{lccc}
                    \toprule
                    Model & $\uparrow$ ANLS (on test) & $\downarrow$ Parameters & $\downarrow$ FLOPs  \\
                    \midrule
                    Teacher: Donut-base    & \textbf{66\%} & 64M (100\%) & 1.57T (100\%) \\
                        \midrule
                        \multicolumn{4}{c}{Coarse-grained students} \\
                        \midrule
                        Donut-T1    & 23\% & 24M (37\%) & 766G (49\%) \\
                        Donut-T2    & 22\% & 24M (37\%) & 766G (49\%) \\
                        Donut-T3    & 28\% & 24M (37\%)  & 766G (49\%) \\
                        Donut-MINT$_{31\%}$ (ours)   & \textbf{50\%} & 20M (31\%) & 757G (48\%) \\
                        \midrule
                        \multicolumn{4}{c}{Fine-grained students} \\
                        \midrule
                    Unstructured-Pruning \cite{sun2024simpleeffectivepruningapproach}  & 46\% & 4.48M (7\%) & 1.57T (100\%) \\
                    Structured-Pruning \cite{zhang2024finercutfinergrainedinterpretablelayer}  & 50\% & 4.48M (7\%) & 188G (12\%) \\
                    Donut-MINT$_{7\%}$ (ours)    & \textbf{51\%} & 4.48M (7\%) & 211G (13\%) \\
                    \bottomrule
                \end{tabular}
            \end{sc}
        \end{small}
    \end{center}
    \vskip -0.1in
\end{table*}

We note that matching parameter count and FLOPs exactly across methods is infeasible, as these quantities do not scale linearly; pruning structural components such as heads or blocks impacts them differently. Still, the FLOPs remain close enough for a meaningful comparison. Notably, unstructured pruning achieves parameter reduction but no FLOPs savings, as it retains dense computation unless specialized sparse inference is used.

These results confirm that MI-guided pruning retains essential computational pathways more effectively than heuristic-based methods, especially under tight resource constraints. Our approach offers a principled alternative to trial-and-error search or hand-designed heuristics, delivering compact models with better performance preservation on Document Visual Question Answering.

\section{Conclusion} \label{sec: conclusion}

In this work, we introduced a novel approach to model compression for document Visual Question Answering by leveraging mechanistic interpretability to guide pruning and architectural simplification, followed by knowledge distillation to recover performance. Rather than relying on generic or heuristic pruning, we analyzed the inner workings of a vision-language model (Donut) to identify components essential to task performance. This yielded DonutMINT, a lightweight student model that retains high accuracy while significantly reducing computational cost and parameter count. Across both coarse and fine-grained regimes, DonutMINT consistently outperformed state-of-the-art pruning baselines under comparable resource constraints.

Our analysis led to the removal of several self and cross attention heads. However, this does not mean these heads are universally redundant; their utility may depend on the task or domain. While mechanistic interpretability offers a fine-grained view of internal functions, it is not yet exploited during distillation, which uses only final logits. Future work could incorporate intermediate activations, now better understood, to define auxiliary losses and enhance distillation beyond conventional methods.

Our current pipeline relies on manual inspection and bespoke tools, but there is considerable potential in automating interpretability-driven pruning. Automation would enable scaling to larger models and diverse architectures, making principled compression more reproducible and widely applicable. The Visually rich Document Understanding (VrDU) setting poses specific challenges, due to complex multimodal interactions that resist manual probing. This motivates the need for automated corruption techniques capable of perturbing latent states in a controlled and semantically meaningful way. Promising directions include structured textual corruption \cite{NEURIPS2020_92650b2e,wang2022interpretabilitywildcircuitindirect} and multimodal interventions \cite{golovanevsky2025vlmsnoticemechanisticinterpretability}, though the semi-structured nature of VrDU inputs calls for more adaptive methods. Progress in automated circuit discovery \cite{conmy2023automatedcircuitdiscoverymechanistic} and efficient attribution-based patching \cite{nanda2023attributionpatching,syed2023attributionpatchingoutperformsautomated} suggests that scalable interpretability pipelines are increasingly feasible. Beyond compression, such tools could support model editing, interpretability audits, and safety checks in high-stakes domains.

\begin{credits}
\subsubsection{\discintname}
The authors have no competing interests to declare that are
relevant to the content of this article.
\end{credits}
%
% ---- Bibliography ----
%
% BibTeX users should specify bibliography style 'splncs04'.
% References will then be sorted and formatted in the correct style.
%
\bibliographystyle{splncs04}
\bibliography{paper}

\end{document}